# Segmentation of Microscopy Data for finding Nuclei in Divergent Images

Shivam Singh
Computer Science Department
Dr. A.P.J. Abdul Kalam Technical University
Lucknow, India
15cs038@aith.ac.in

Stuti Pathak
Computer Science Department
Dr. A.P.J. Abdul Kalam Technical University
Lucknow, India
16cs050@aith.ac.in

## ABSTRACT

Every year millions of people die due to disease of Cancer. Due to its invasive nature it is very complex to cure even in primary stages. Hence, only method to survive this disease completely is via forecasting by analyzing the early mutation in cells of the patient biopsy. Cell Segmentation can be used to find cell which have left their nuclei. This enables faster cure and high rate of survival. Cell counting is a hard, yet tedious task that would greatly benefit from automation. To accomplish this task, segmentation of cells need to be accurate. In this paper, we have improved the learning of training data by our network. It can annotate precise masks on test data. we examine the strength of activation functions in medical image segmentation task by improving learning rates by our proposed *Carving Technique*. Identifying the cells' nuclei is the starting point for most analyses, identifying nuclei allows researchers to identify each individual cell in a sample, and by measuring how cells react to various treatments, the researcher can understand the underlying biological processes at work. Experimental results shows the efficiency of the proposed work.

## 1. Introduction

Cell segmentation and counting can be a laborious task that can take up valuable time from research. Typically, a scientist must manually estimates the number of cells in a local grid within an image [5]. This is repeated at various grid points across the plate to get a mean density which is then used for estimating the total number of cells. These density based techniques suffer from several drawbacks: first, they require a human to manually count the number of cells, introducing the possibility of subjective errors; second, they require a significant amount of time commitment, which could be better used for understanding, designing, and performing a new series of experiments; finally, it is not completely obvious how error bars can be obtained from such an analysis. Although more sophisticated tools do exist for this task, they can be costly, rely on closed-source software, and do not address the issue of quantifying error**.** Segmenting lesions or abnormalities in medical images demands a higher level of accuracy than what is desired in natural images. While a precise segmentation mask may not be critical in natural images, even marginal segmentation errors in medical images can lead to poor user experience in clinical settings [6]. For example, an erroneous measurement of nodule growth in longitudinal studies can result in the assignment of an incorrect Lung-RADS category to a screening patient. It is therefore desired to devise more effective image segmentation architectures that can effectively recover the fine details of the target objects in medical images.

The contributions of this work can be summarized as follows:
- To address the need for more accurate segmentation in medical images, we present *Carving Technique* in standard U-Net architecture. This process increases the Intersection over Union (IoU), reduces validation loss and exhibits a better performance than a traditional U-net.

- The underlying hypothesis behind our architecture is that the model can more effectively capture fine-grained details of the foreground objects when high-resolution feature maps from the encoder network are gradually enriched prior to fusion with the corresponding semantically rich feature maps from the decoder network.

- In traditional U-Net, which directly fast-forward high-resolution feature maps from the encoder to the decoder network, resulting in the fusion of semantically dissimilar feature maps. According to our experiments, the suggested technique is effective, yielding significant performance gain over U-Net and optimized U-Net.

The rest of the paper is organized as follows: In Section 2, we have mentioned the previous and similar works done earlier in segmentation tasks. In Section 3, we have described our methodology, loss function used to evaluate performance and performance metric to determine quality of segmentation task. Finally, In Section 4, we have provided experimental proofs to verify our methodology and have provided plots, tables and visualizations of segmentation task. Finally the paper is concluded in Section 5.

## 2. Previous Works

Segmentation using CNNs is an important problem in computer vision and significant progress has been made over the past few years [7, 10, 11]. We note that cell segmentation using CNNs has been addressed at least once before in the literature. Van Valen, et al. developed DeepCell, which treats the segmentation task as classification problem on a pixel-by-pixel basis [16]. While successful at classifying different cell types in images, DeepCell produces fairly low-resolution segmentation masks and does not aim to solve the cell counting problem. Our work however target to improve the segmentation to detect discrete cell by carving the separation boundaries in our output by learning them precisely. By accurate separation of cells in out output it makes us easy to automate cell counting.

To make our network performs better, different activation functions are applied in our network. Activation function is an important component of a neural network. The activation layers adds non-linearity to the weights so that the network can deal with more complex tasks. Previously, sigmoid [8] has been used as the activation function. However, sigmoid will saturate during training. Next, ReLU [9] has been proposed to solve the saturation issue. While using ReLU as the activation function, the learning rate should be carefully adjusted because ReLU gets saturated in negative axis, and a big learning rate will "kill" some neurons. To address this efficiency, Exponential ReLU (ELU) [12] has been proposed. ELU does not get saturated in the negative axis immediately. However, saturation still happens when the network gets deeper. When ELU gets saturated, it is no different from ReLU. In this paper, we use ELU and ReLU simultaneously. We want to show the underutilized capability of activation functions.

## 3. METHODOLOGY

In our proposed *Carving Technique,* we have exploited the fact that by combining activation function be can further improve the performance of our deep learning architectures. Activation functions are conventionally used as an activation function for the hidden layers in a deep neural network [14]. The motivation behind our technique is the Multi-Armed Bandit Problem from domain of reinforcement learning [13]. In which we want to optimize the reward by not getting trapped in sub-optimal region. We have proved our hypothesis by the set of experiments. We observed that single activation functions get trapped in suboptimal regions too early. In our proposed work, we don't allow gradients to saturate too early. Hence, it gives a greater room for reduction of losses and greater performance in task of segmentation.

### 3.1 Model Setup

We have used standard U-net architecture consists of a down-convolutional part and up-convolutional part [3]. The down-convolutional part aims at extracting features for classifying each voxel into one or zero. It consists of repeated application of two $3 \times 3$ convolutions. At each downsampling step the number of feature channels is doubled. And the up-convolutional part aims at locating regions of interest (ROI) more precisely. Every step in up-convolutional part consists of an upsampling of the feature map followed by a $2 \times 2$ convolution that halves the number of feature channels, a concatenation with the correspondingly cropped feature map from the contracting path, and two $3 \times 3$ convolutions. We have used simultaneous ELU and ReLU respectively in encoder segment of this network architecture which gives us better performance by not trapping in sub-optimal region due to the proposed *Carving Effect,* In artificial neural networks, the activation function play an important role. Rectifier liner unit (ReLU) is the most popular activation function for deep neural network [15]. Exponential Liner Unit (ELU) replace the negative part in ReLU with exponential function, which is helpful to make the average of output close to zero [12]. In our network, we initially use ELU with all layers. Both ReLU and ELU are widely used in segmentation task. In this work, we find the combination of ReLU and ELU can improve the segmentation performance. Neural networks usually suffer low coverage rate because of vanishing gradient, especially for deep network. ELU provides a buffer in negative axis so that it will not saturate immediately. However, ELU still suffers the saturation problem when network gets deeper. When ELU saturated to negative values, there will be no difference between ELU and ReLU. To overcome this issue, we have proposed a method using ELU and ReLU simultaneously. Since ELU will be saturated when the network going deeper and has the same effect as ReLU, we can simply use ReLU in layers of decoder phase i.e. layers $6^{th}$, $7^{th}$, $8^{th}$ and $9^{th}$. The saturated negative values in the following ELU layers will be reset to 0 because of ReLU, which means these following layers are not saturated anymore. Therefore, model does not early stops in sub-optimal region.

Figure 1 describes our Network with simultaneous use of ELU and ReLU in encoder segment where only ReLu in decode segment of Standard U-Net architecture.

Segmentation of Microscopy Data for finding Nuclei in Divergent Images

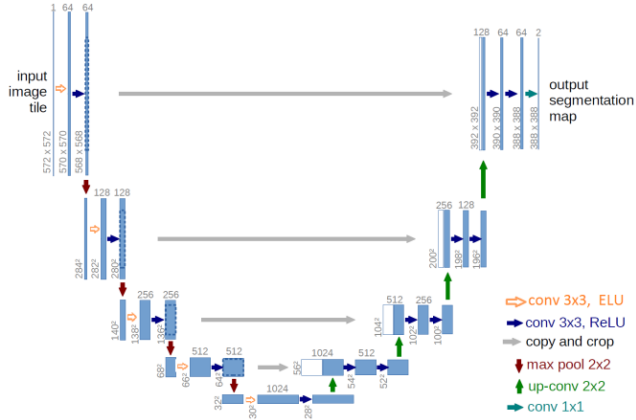

Figure 1: Our Network with simultaneous use of ELU and ReLU in encoder segment and only ReLu in decode segment of Standard U-Net architecture.

### 3.2 Loss Function

We have chosen Binary Cross Entropy (BCE) as loss function in our experiments. The typical cost function that one uses in segmentation task is computed by taking the average of all cross-entropies in the sample. For example, suppose we have N samples with each sample labeled by i=1, 2…., N. The loss function is then given by:

$$BCE = -\frac{1}{N}\sum_{i=0}^{N} y_i \cdot log(\hat{y}_i) + (1 - y_i) \cdot log(1 - \hat{y}_i)$$

Using Binary Cross Entropy, we calculate the validation loss of various models for comparison.

### 3.3 Performance Metric

The performance of our models is evaluated on the mean average precision at different Intersection over Union (IoU) thresholds. The IoU of a proposed set of object pixels and a set of true object pixels is calculated as:

$$\text{IoU}(A,B) = \frac{A \cap B}{A \cup B}$$

The metric sweeps over a range of IoU thresholds, at each point calculating an average precision value. The threshold values range from 0.5 to 0.95 with a step size of 0.05, i.e. (0.5, 0.55, 0.6, 0.65, 0.7, 0.75, 0.8, 0.85, 0.9, 0.95). In other words, at a threshold of 0.5, a predicted object is considered a "hit" if its intersection over union with a ground truth object is greater than 0.5. We aim at getting a higher value for IoU to achieve accurate segmentation.

### 4. Experimental Evaluation

In this section, we provide the experimental results of our model. We have compared three models, each one of which is a variation of U-Net. The first model is the proposed model which employs the *Carving Technique*. It consists of ELU and ReLU in encoder phase and ReLU in decoder phase. The second model is Optimized U-Net which consist all activations as ELU. The third model is baseline standard U-Net with all activations as ReLU.

### 4.1 Dataset

We have used the dataset provided in Kaggle Data Science Bowl Challenge 2018 for evaluation of our models [4]. It consists of microscopy images of a large number of segmented nuclei images. The images were acquired under a variety of conditions and vary in the cell type, magnification, and imaging modality (brightfield vs. fluorescence). Training set contain 670 images with their corresponding masks and the test set contain of 65 images.

### 4.2 Implementation Details

Keras [17] with Google TensorFlow [18] backend was used to implement the deep learning algorithms in this study, with the aid of other scientific computing libraries: matplotlib, numpy, and scikit-learn [19].

### 4.3 Results and Discussion

In our experiments, we have run all three of our models for 35 epochs with batch side 8 and patience is set to 7 i.e. model early stops if losses doesn't improve for 7 epochs [17]. We have repeated our experiment 10 times we the keeping hyperparmters fixed as value mentioned above. Finally, we have recorded the mean of all 10 runs for each model and mentioned them in Table 1. It can be seen clearly that by using single activation function both of the models, i.e. baseline U-Net with only ReLU activation layers and Optimized U-Net with only ELU get trapped in sub-optimal range. Therefore, model is not learning the training data which leads to poor segmentation.

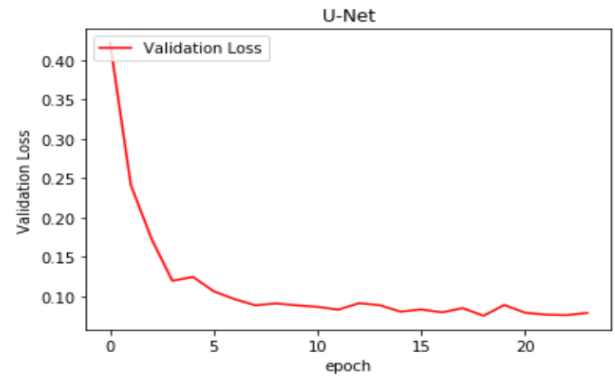

Figure 2.a: The figure shows Validation loss of baseline U-Net model.

In Figure 2.a, the validation loss of the baseline U-Net performs early stopping by getting trapped in sub-optimal region is depicted.

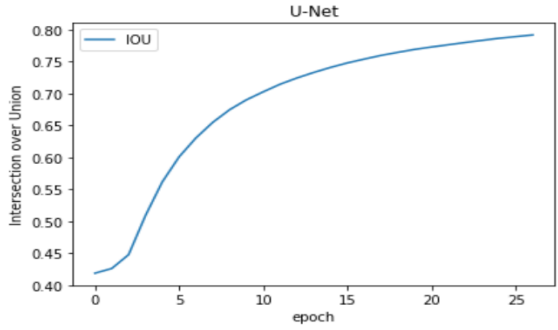

Figure 2.b: The figure shows IoU of baseline U-Net model.

Figure 2.b illustrates that due to early stopping of model the IoU also increases up to a extent even though further improvements are possible. Hence, due to poor IoU less accuracy is obtain is segmented images.

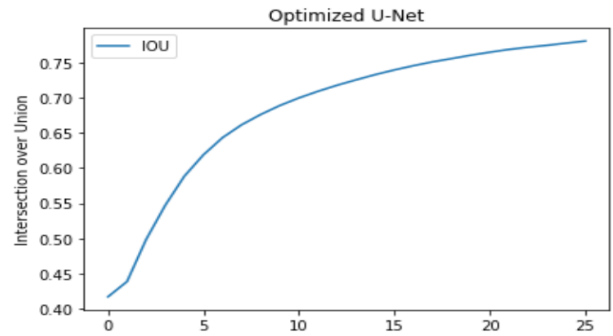

Figure 3.b: The figure illustrates IoU of Optimized U-Net

Figure 3.b depicts that due to early stopping of model the IoU also increases up to a extent even though further improvements are possible. Hence, due to poor IoU less accuracy is obtain is segmented images.

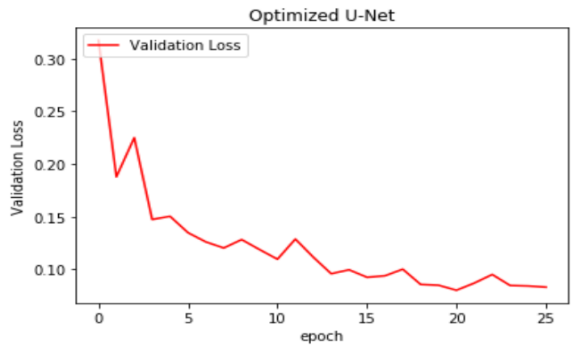

Figure 3.a: The figure shows Validation loss of Optimized U-Net model which early stops at 26 epoch.

Figure 3.a shows the validation loss of the optimized U-Net in this model where we have replaced all the ReLU activation layers with ELU as it gives better performance than ReLU [12]. But unfortunately, it also performs early stopping by getting trapped in sub-optimal region.

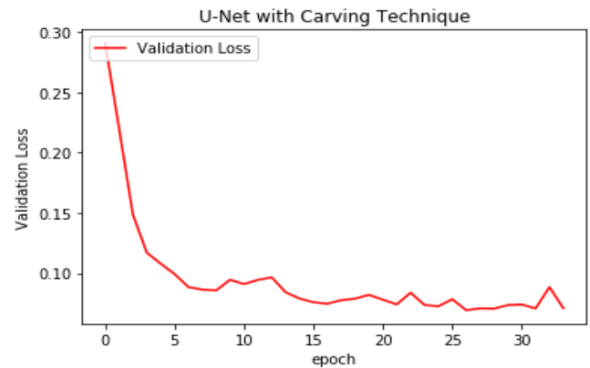

Figure 4.a: The figure shows Validation loss of U-Net using our proposed carving technique.

Figure 4 describes the validation loss of U-Net Model using our proposed *Carving Technique*. Hence, model does not early stops and neither get trapped in sub-optimal range. Due to this reason, model continues to learn new features over all the epochs.

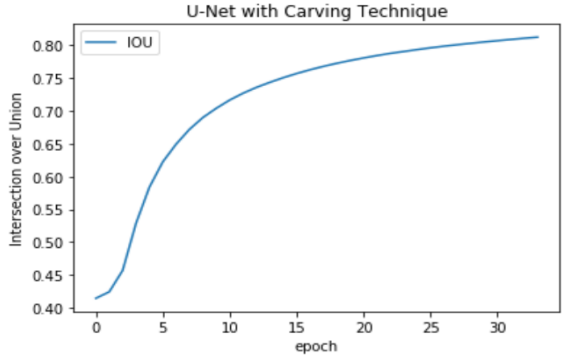

Figure 4.b: The figure shows IoU of U-Net using our proposed carving technique.

In Figure 4.b, the model with our proposed *Carving Technique* doesn't early stops because use of simultaneous ELU and ReLU which doesn't allow ELU to saturate early. Hence, it prevents it to getting trapped in sub-optimal regions of validation loss i.e. why IoU also keeps increasing gradually. As a result we have obtain precise separation boundaries in our task of cell segmentation.

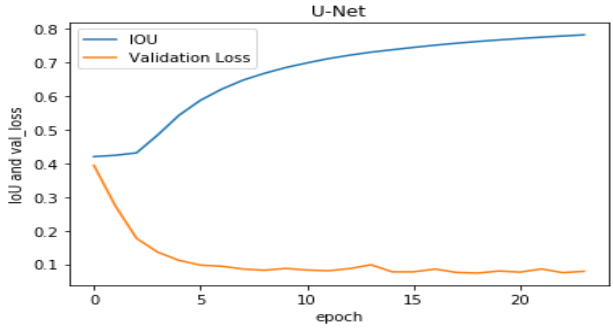

(a)

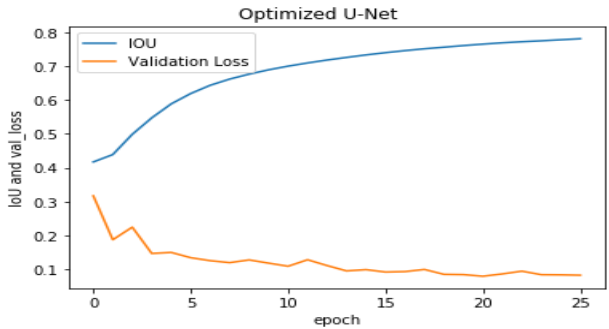

(b)

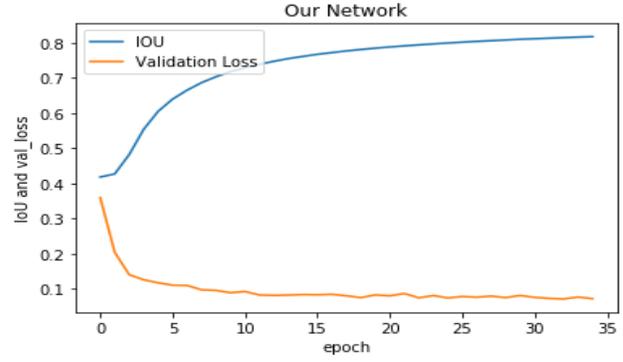

(c)

Figure 5: The figure illustrates comparison between three model over performance metric and validation loss.

In Figure 5, we have provided three plots corresponding to our three models i.e. (a) baseline U-Net, (b) Optimized U-Net and (c) Our Network using *Carving Technique*. It is clear from the results that the experimental results prove our initial hypothesis.

Table 1: Performance of models with different Activation Layers

| Model | Mean IoU | Validation Loss | Early Stops in Epoch |
|---|---|---|---|
| U-Net | 78.2% | 0.082 | 22 |
| Optimized U-Net | 81.3% | 0.076 | 24 |
| U-Net with Carving Technique | **88.62%** | **0.068** | Doesn't Early Stop |

Table 1 shows the quantitative comparison between the model with proposed method with other models for the task of segmentation.

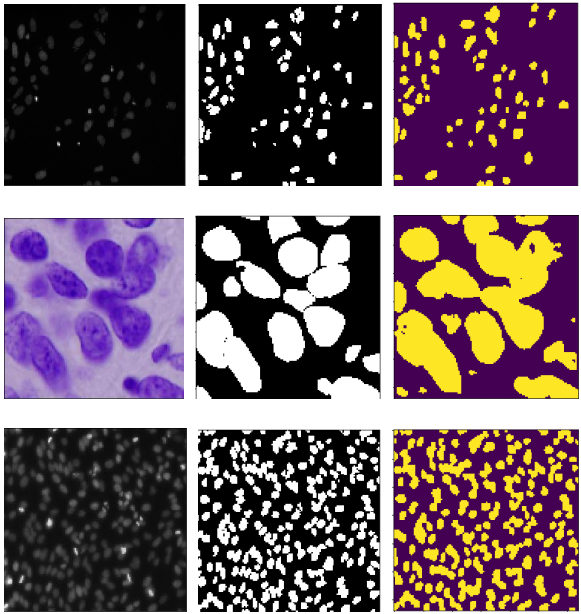

Figure 6: The segmentation is performed by the model with carving effect on images with high degree of overlapping.

In figure 6, it is clear that our model shown in figure 1 fits well on training dataset. Hence, we get more accurate separation boundaries in the annotations produced by our model. Above, shown pictures contain the cell with high degree of overlapping. Model which employed our technique performs significantly well on these type of images.

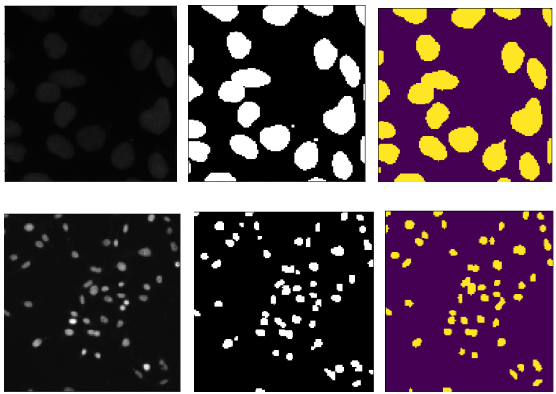

Figure 7: The segmentation is performed by the model with carving effect on images with relatively isolated cells.

Figure 7 shows that our model performs significantly on images with relatively isolated images generating accurate cell boundaries in masks produced by our model.

## 5. Conclusion

In this paper, we proved that using ELU and ReLU simultaneously gives us better results in image segmentation task rather than using only one of them. This method also yields better performance in other image classification tasks also.

We are looking ahead to generalize the method in future. In this paper, we have proved the results by experimentation. But a better formal mathematical explanation can be derived. We are working on proving this method by inferring facts from [12] and [14]. We can use a reinforcement learning approach to exploit potential of existing activation functions.